\documentclass{article}

\usepackage[numbers]{natbib}
\usepackage{graphicx}
\usepackage{subfig}
\usepackage[final]{neurips_2024}

\usepackage[utf8]{inputenc} 
\usepackage[T1]{fontenc}    
\usepackage{hyperref}       
\usepackage{url}            
\usepackage{booktabs}       
\usepackage{amsfonts}       
\usepackage{nicefrac}       
\usepackage{microtype}      
\usepackage{xcolor}         

\title{Uncovering Uncertainty in Transformer Inference}

\author{
  Greyson Brothers \\
  Johns Hopkins University \\
  Applied Physics Laboratory \\ 
  \texttt{greyson.brothers@jhuapl.edu} \\
  \And
  Willa Mannering \\
  Johns Hopkins University \\
  Applied Physics Laboratory \\ 
  \texttt{willa.mannering@jhuapl.edu} \\
  \AND
  Amber Tien \\
  Johns Hopkins University \\
  Applied Physics Laboratory \\ 
  \texttt{amber.tien@jhuapl.edu} \\
  \And
  John Winder \\
  Johns Hopkins University \\
  Applied Physics Laboratory \\ 
  \texttt{john.winder@jhuapl.edu} \\
}
\begin{document}

\maketitle

\begin{abstract}
We explore the Iterative Inference Hypothesis (IIH) within the context of transformer-based language models, aiming to understand how a model's latent representations are progressively refined and whether observable differences are present between correct and incorrect generations. Our findings provide empirical support for the IIH, showing that the \(n^{th}\) token embedding in the residual stream follows a trajectory of decreasing loss. Additionally, we observe that the rate at which residual embeddings converge to a stable output representation reflects uncertainty in the token generation process. Finally, we introduce a method utilizing cross-entropy to detect this uncertainty and demonstrate its potential to distinguish between correct and incorrect token generations on a dataset of idioms. 
\end{abstract}

\section{Introduction and Related Work}
Transformer-based architectures \cite{vaswani2017attention} currently dominate artificial intelligence applications and serve as the underlying architecture for most Large Language Models (LLMs). While LLMs show impressive emergent abilities, these models exhibit limitations such as hallucinations and biased outputs which pose significant societal challenges \cite{park2023ai}. Inaccurate outputs can mislead users, while malicious actors can exploit AI models to create deceptive images, videos, and text to represent fictional occurrences as truth. Mitigating harms caused by model misuse, biased outputs, or misalignment with human values is a primary motivation behind research and policy decisions related to AI interpretability \cite{down2021proposal}. 

In this work we investigate a novel method for detecting uncertainty during the token generation process of transformer-based language models. One framework that has emerged for understanding the feed-forward behavior of residual models, such as the transformer, is the Iterative Inference Hypothesis (IIH) \cite{dar2022analyzing,greff2016highway,lad2024robustness}. This hypothesis posits that predictions are formed in the residual stream, and that each block in a residual architecture incrementally updates these predictions in a direction of decreasing loss \cite{jastrzkebski2017residual}. A related line of research on in-context learning has suggested that transformers trained on autoregressive tasks are closely related to formulations of iterative optimization algorithms, namely gradient descent \cite{vonoswald2023iclGradient}. 

We combine these two threads of research, framing transformer inference as an optimization process that iteratively updates the $n^{th}$ input embedding (i.e. the last word in the input sequence) to converge toward the most likely next-token embedding given the context and model weights. We propose methods for evaluating how input embeddings evolve towards output embeddings and find preliminary evidence of observable differences between correct and incorrect outputs, suggesting that these metrics could serve as useful indicators of a model's output certainty. Our contributions are as follows: 
\begin{enumerate}
\item[1.] We find preliminary evidence that the $n^{th}$ token embedding in the residual stream follows a path of decreasing loss in token embedding space, supporting the IIH. 
\item[2.] We propose a novel method for detection of uncertainty during the token generation process and find preliminary evidence that this metric reflects observable differences in correct and incorrect generations. 
\end{enumerate}

\section{Methods}
\label{sec:methods}
\begin{figure}[!ht]
  \centering
  \subfloat{\includegraphics[width=0.5\textwidth]{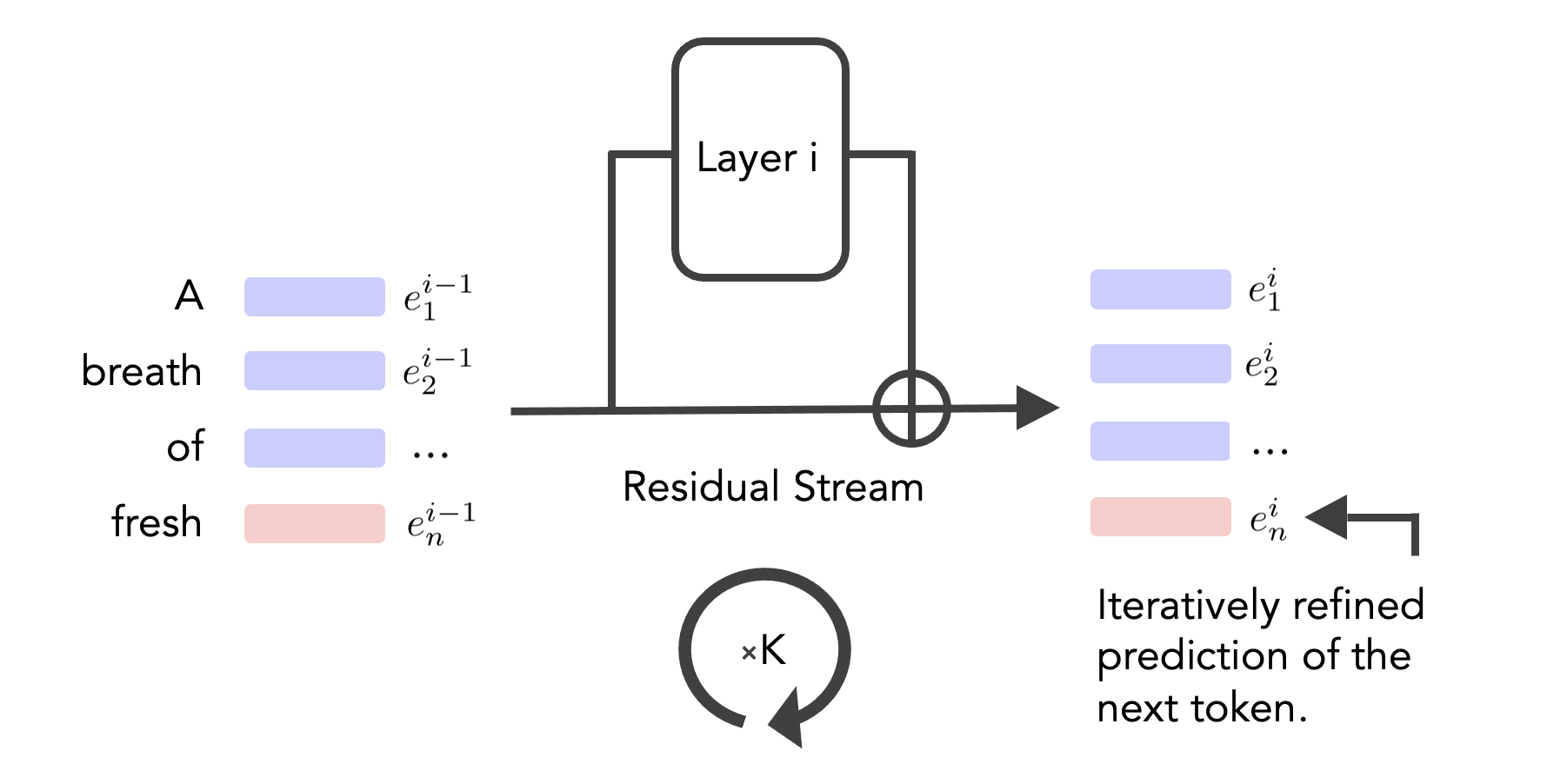}}
  \caption{The transformer as a recurrence relation, iteratively refining a prediction for the next token. }
  \label{transformer}
\end{figure}

Here we define methods that offer insight into how representations in the residual stream evolve during inference. The transformer architecture, excluding token embedding and unembedding, can be succinctly represented by the following recurrence relation: $ \mathbf{r}_{i+1} = \mathbf{r}_i + L_{i+1}(\mathbf{r}_i ),$ where $\mathbf{r}_{i} = [e_1^{i}, ..., e_n^{i}]$ represents the set of token embeddings in the residual stream after an update from layer $L_{i}$. Each layer contains attention and feed-forward sublayers. With $\mathbf{r}_0$ as the set of input token embeddings plus positional encodings, the residual stream is the sequence $(\mathbf{r}_{0}, \mathbf{r}_{1}, ..., \mathbf{r}_{k})$ for a model with $k$ layers. In this work we are particularly interested in tracking the evolution of the embedding of the $n^{th}$ input token, $e_n^i$, as shown in red in Figure \ref{transformer}. Its residual representation after the final layer update, $e_n^k$, is used to predict the next token in an autoregressive framework. 

In the convolutional architecture of the original residual network (ResNet), the residual stream contained shortcut connections that transformed the basis of latent space in order to reduce its dimensionality at regular intervals \cite{he2015residual}. However, in the transformer, the residual stream has an entirely linear structure that preserves the basis of the embedding space \cite{elhage2021mathematical}. Each layer adds an output to each residual embedding, corresponding to translations in token embedding space. Thus the residual stream can be viewed as a path through token embedding space, and each point along the path can be mapped back to a distribution over tokens via methods like logit lens \cite{nostalgebraist2020logitlens}. To get the intermediate distribution predicted by the residual stream after layer $L_i$, we pass $e_n^i$ through the output layer norm and then to the model's output head to obtain logits over the vocabulary. We refer to these logits as residual predictions. This framework forms the basis of our methods and analysis. 

\subsection{Residual Cross-Entropy}
\label{sec:cross-entropy}
We present a method by which cross-entropy can be used to measure the evolution of residual predictions during token generation. The IIH posits that the output of each layer updates the residual prediction in a direction of decreasing loss. We thus examine per-layer changes in cross-entropy, the loss function used to train transformers on autoregressive language tasks \cite{henighan2020scalinglaws}, including the specific model we examine here (GPT-2 XL) \cite{radford2019language}. 

Cross-entropy is a measure of dissimilarity, requiring a candidate and a target distribution. We use the residual predictions described above as the candidate, and for the target distribution we examine two choices: (1) the one-hot distribution of the token sampled deterministically by taking the argmax the model's predicted probabilities, denoted $\hat{y}$, and (2) the one-hot distribution of the ground truth next token given by the dataset, denoted $y$.  These measures are equivalent to the negative log likelihood of the sampled and target tokens respectively, as elaborated in \ref{sec:appendix-cross-entropy}. When the sampled output is correct, or $\hat{y} = y$, these two measures are the same. Examining how layer updates affect cross-entropy with respect to the training objective $y$ is necessary for our evaluation of the IIH, but the ground truth next token may be unavailable or ambiguous at inference time, making $\hat{y}$ more practical for analyzing the residual stream in an online setting. Some existing works have examined  Kullback-Leibler divergence between residual predictions and the final logits output by the model \cite{nostalgebraist2020logitlens, lad2024robustness}, rather than $\hat{y}$ as we do here. Our approach implicitly penalizes generations that have high output entropy, which is advantageous for our study of uncertainty during generation. We provide a comparison between the two in \ref{sec:appendix-output-targets}.

\subsection{Model} 
To investigate how representations evolve in the residual stream we examine GPT-2 XL \cite{radford2019language}, which has 48 layers and 1.5 billion parameters. We chose this model as it has low compute requirements while still being representative of frontier model architectures. In addition, it has widely used open source implementations and official weights made publicly available by OpenAI on HuggingFace\footnote{Official weights for GPT-2 XL hosted on Hugging Face: https://huggingface.co/openai-community/gpt2-xl}.

\subsection{Inference Data}
For our preliminary study of the IIH and model uncertainty, we chose a simple dataset with a wide variety of generation difficulty: English idiom completion. An idiom completion task provides a relatively clear-cut, single token "correct" and "incorrect" answer, making it straightforward to evaluate. This dataset is intentionally challenging, where some answers would be hard or nearly impossible to guess given the input context or brevity of the idiom. 

The idiom dataset consists of 330 static idioms taken from the EPIE Dataset \cite{saxena2020epie}. To build our dataset, we split each static idiom so that the final word serves as the "correct" output for the model.  To guide the model in completing the idiom, we added instructions to the start of the idiom phrase. The instructions read: "The following prompt is the beginning of a popular English idiom, please respond with a single word to complete the phrase." Thus, each prompt in this dataset consists of the instructions + the first words of an idiom. We excluded 29 idioms from the EPIE dataset because the target outputs were represented by more than one token in the vocabulary, such as ["help", "ful"].

\section{Results}

In this section, we examine the evolution of residual representations in GPT-2 XL on the idiom inference dataset by analyzing the per-layer change in cross-entropy. Our main objectives are, first, to evaluate whether there is evidence supporting the IIH, and second, to determine if these metrics reveal a noticeable difference between correct and incorrect output distributions that could aid in developing a measure of uncertainty for a model's predictions.  

\begin{figure}[!ht]
  \centering
  \subfloat[][Cross-Entropy (\textit{layer}, $\hat{y}$)]{\includegraphics[width=.485\textwidth]{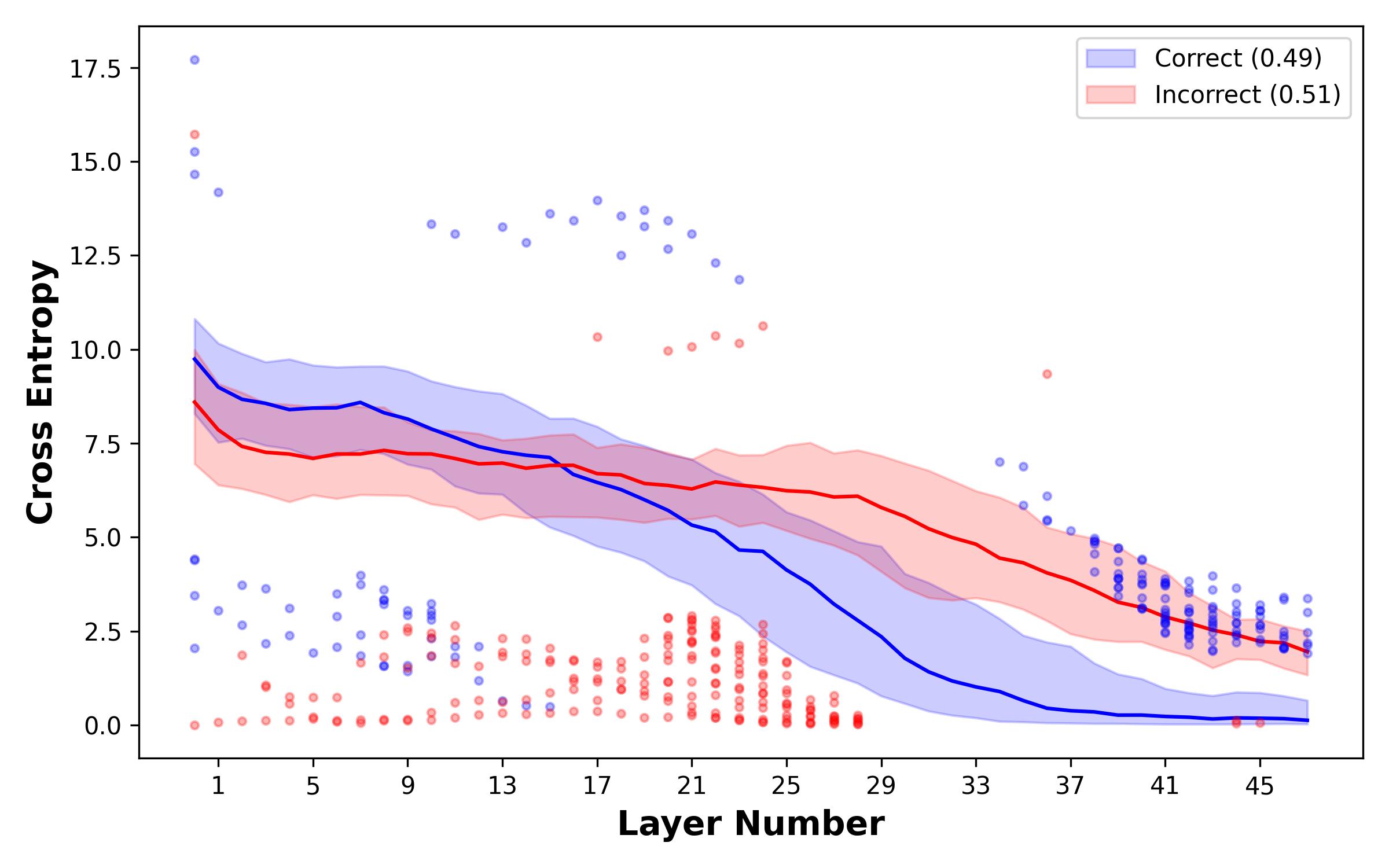}}\quad
  \subfloat[][Cross-Entropy (\textit{layer}, $y$)]{\includegraphics[width=.485\textwidth]{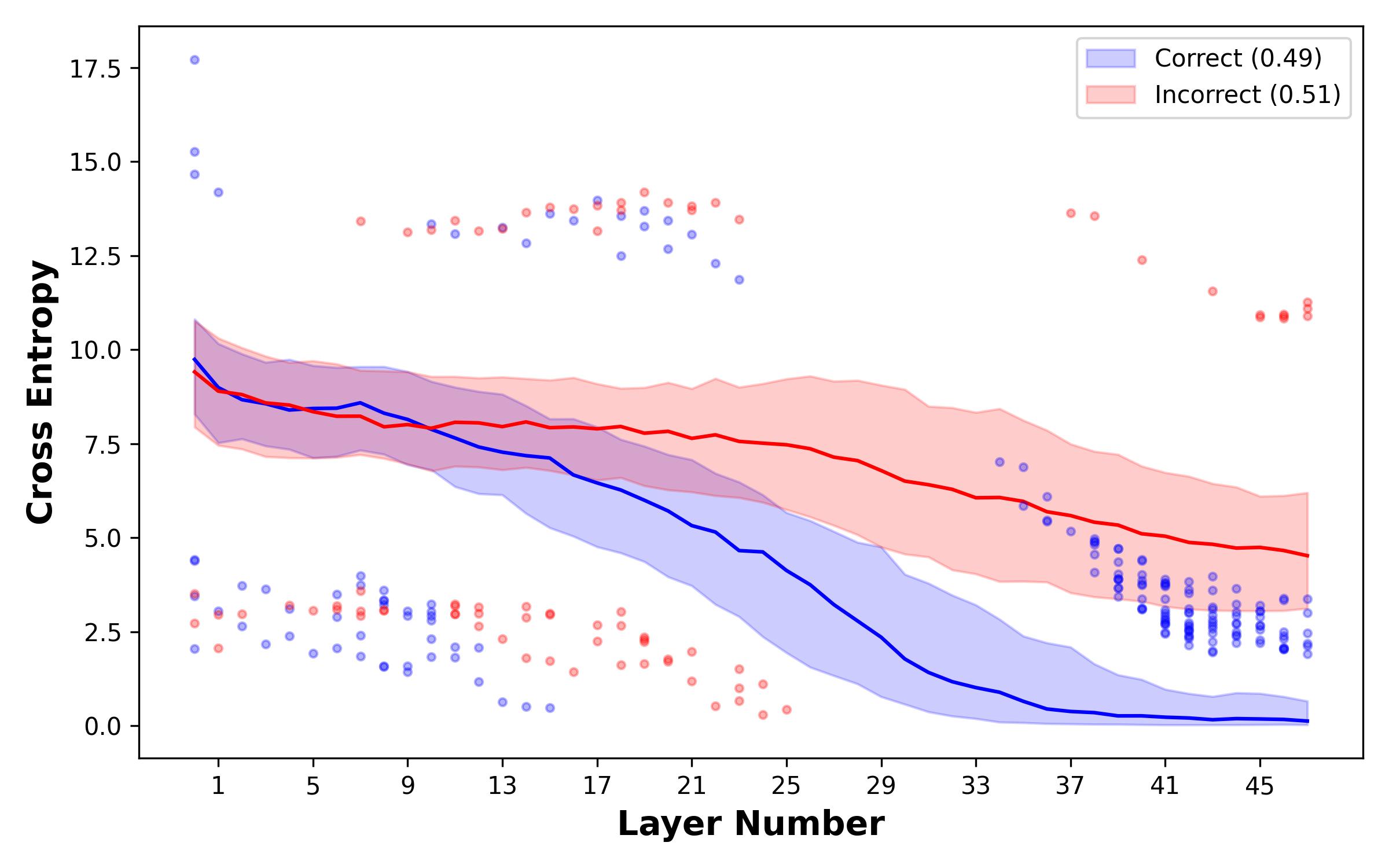}}  
  \caption{Plots showing the cross-entropy between the residual prediction at each layer and a target distribution. The median, inter-quartile ranges, and outliers of correct and incorrect generations are plotted for 330 samples. (Left) The token predicted by the model $\hat{y}$ is used as the target. (Right) The ground-truth token $y$ from the dataset is used as the target.  }

  \label{fig:sub1}
\label{idiom-lineplot}
\end{figure}

The cross-entropy per model layer for the idiom dataset is presented in Figure~\ref{idiom-lineplot}. To generate these results, we first feed a prompt into the model, recording the \(n^{th}\) residual embedding before and after the update from each layer. We then calculate the cross-entropy between these residual predictions and a target. For Figure~\ref{idiom-lineplot}a we take the target to be the token predicted by the model, $\hat{y}$, as described in Section~\ref{sec:cross-entropy}. For Figure~\ref{idiom-lineplot}b, we let the target be the ground-truth next token, $y$, from the idiom dataset. Distributions of correct generations ($\hat{y}=y$) for each figure are plotted in blue and incorrect generations are plotted in red. This approach allows us to clearly observe the evolution of representations that ultimately lead to correct predictions during the inference process. 

In other words, Figure~\ref{idiom-lineplot}a displays how the embeddings in the residual stream evolve towards an arbitrary output representation, while Figure~\ref{idiom-lineplot}b shows how it evolves towards the most likely next token according to the dataset. A distinct separation is observable between the correct and incorrect distributions for both cross-entropy plots, suggesting these measures may be useful for understanding the certainty of a model's output as it is being generated. The separation is more pronounced when the target is the ground truth, as a result of the incorrect generations failing to converge to the correct representation in embedding space. Figure~\ref{idiom-lineplot}b demonstrates clear evidence for the IIH, with the median layer update decreasing loss with respect to the ground truth nearly monotonically throughout the model for both correct and incorrect generations. We provide a table with the average decrease in cross-entropy per layer for (b) in \ref{sec:delta-loss}.

\begin{figure}[h]
\centering
\subfloat[Cross-Entropy(\textit{output}, $\hat{y}$)]{\includegraphics[scale=0.42]{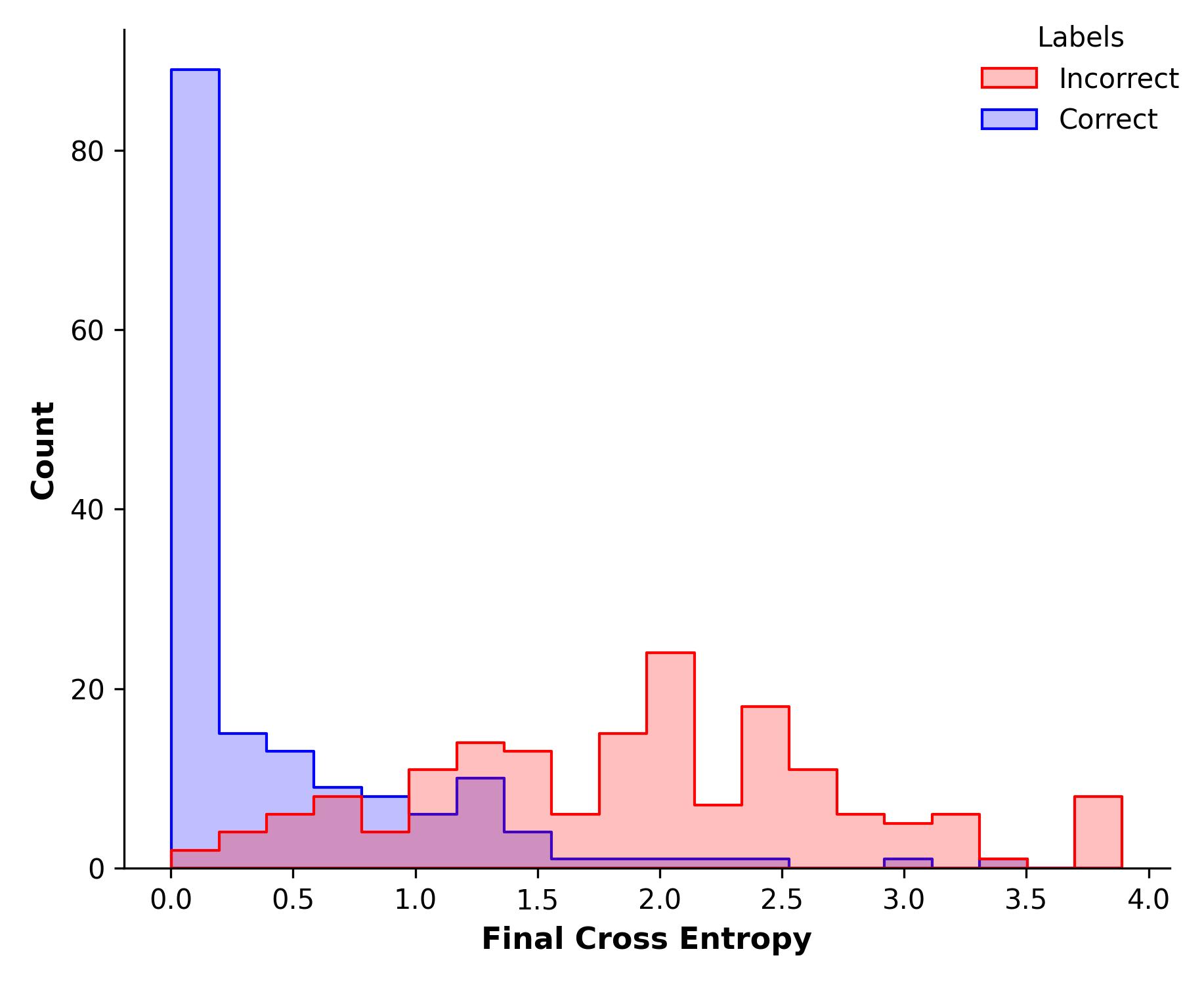}}
\hspace{0.07\textwidth}
\subfloat[Output Cross-Entropy ROC Curve]{\includegraphics[scale=0.35]{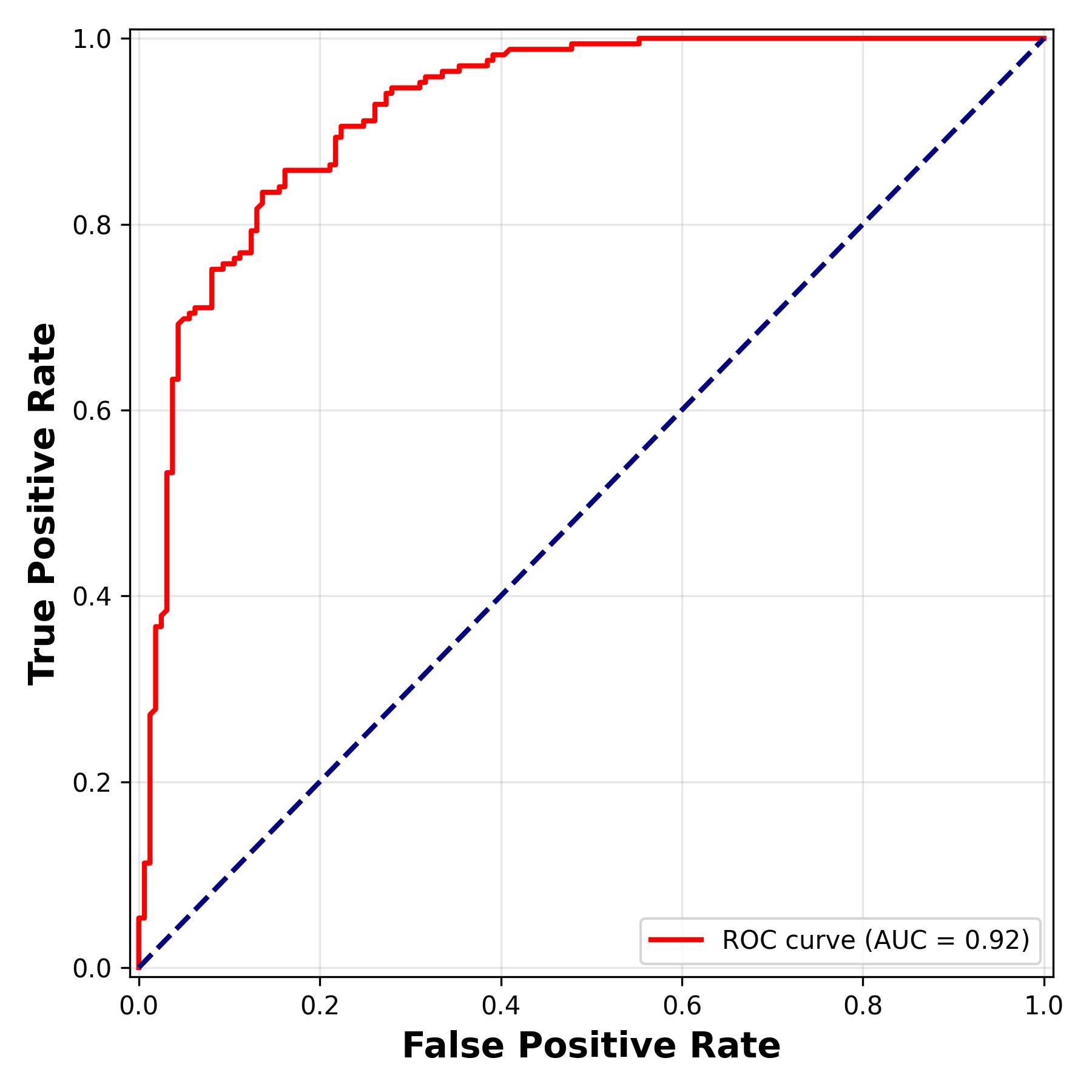}}
\caption{ (Left) Distributions of correct and incorrect generations according to final layer cross-entropy with target $\hat{y}$. (Right) The corresponding ROC curve. As indicated by the AUC of 0.92, the output cross-entropy is a strong predictor of correct generations for the idiom dataset. }
\label{idiom-output-cross-entropy}
\end{figure}

In Figure~\ref{idiom-output-cross-entropy}a we show the cross-entropy distributions of the model's output logits with respect to $\hat{y}$, corresponding to the vertical slice over the last layer in Figure~\ref{idiom-lineplot}a. We observe that the distribution of correct generations is exponential with a mean of $0.43$, indicating that correct generations tend to converge more closely to a one-hot distribution and thus implying their output distributions have lower entropy. The incorrect predictions are normally distributed with a mean of $1.91$ and a standard deviation of $0.86$, indicating that final predictions tend to be further in distribution from $\hat{y}$ and thus have higher entropy. By visual inspection, any generations resulting in a output cross-entropy greater than $1.5$ are very likely to be incorrect for this dataset. In Figure~\ref{idiom-output-cross-entropy}b we plot a receiver operating characteristic (ROC) curve for the output cross-entropy and observe an area under the curve (AUC) of $0.9239$, indicating that output cross-entropy is a strong predictor of correct vs incorrect generation on the idiom dataset. This was corroborated with a Mann-Whitney $U$ test yielding a $\rho$ statistic with the same value. 

\begin{figure}[!ht]
  \centering
  \subfloat{\includegraphics[width=1.0\textwidth]{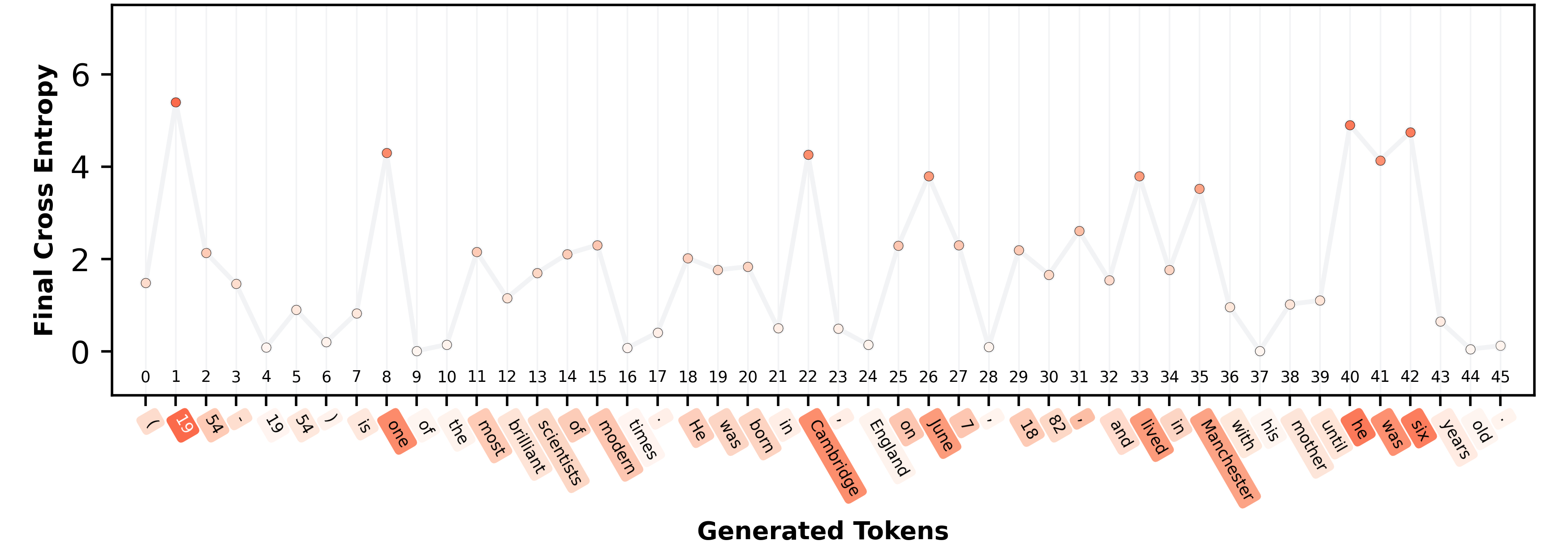}}
  \caption{Output cross-entropy per generated token given the open-ended prompt "Alan Turing". }
\label{open-ended-generation}
\end{figure}

A potential application of this metric is visualized in Figure~\ref{open-ended-generation}, measuring the output cross-entropy per token on an open ended generation task. Again, we measure the cross-entropy distributions of the model's output logits with respect to $\hat{y}$ since the model has access to this information at inference time. This generation was produced using a seed of 42, a temperature of 0.8, and the prompt "Alan Turing". 

In response, GPT-2 XL generates a number of erroneous historical facts, each of which is observed to have a spike in cross-entropy relative to the rest of the sequence. For reference, Alan Turing was born in 1912 in London, England, attended King's College in Cambridge, and died on June 7, 1954. The cross-entropy values are lowest when the language heavily implies the next token, such as with generated tokens 9, 10, 16, 24, 37, and 44. Here the space of possible next tokens is heavily constrained, in contrast to open-ended portions of the sequence encountered when generating tokens 8, 22, 33, 40, and 41. Large cross-entropy values in this sequence are also observed on for tokens representing dates, locations, and other facts - see tokens 1, 22, 26, 35, and 42. 

This example concisely illustrates trends observed over numerous generation studies, but it alone is not sufficient for general claims. Anecdotally, the cross-entropy measure appears to capture both uncertainty inherent in the prompt and the uncertainty of the model. It appears to be insufficient to disambiguate between these two sources of uncertainty. Additionally, if a model is confidently incorrect, this metric would not capture any uncertainty to indicate a potential mistake $-$ a limitation humans face as well. 

Output cross-entropy can be computed cheaply at inference time and these results indicate that it can potentially be used by a naïve classifier to indicate the likelihood of an incorrect generation. Future work will examine if this result holds across models and datasets. In the appendix \ref{sec:appendix-regret-samples}, we present a table showing the intermediate predictions for the highest and lowest cross-entropy results in Figure~\ref{idiom-output-cross-entropy}a. These samples further show that output cross-entropy appears to correspond with prompt open-endedness and the model's uncertainty given the prompt. These combined results indicate that the correctness and perhaps amount of certainty for the next generated token can be measured by the rate and degree to which embeddings converge to stable output representations in the residual stream of transformers.  

\section{Conclusion}
In this work we investigate the Iterative Inference Hypothesis as applied to the transformer architecture on autoregressive language modeling problems. We provide a mechanism for investigating the evolution of predictions in the residual stream and find empirical evidence to support the hypothesis. In addition, we propose a novel method for observing uncertainty during the token generation process by measuring the cross-entropy between the model output logits and a one-hot distribution representing the deterministically sampled token $\hat{y}$. Using this metric we find distinct differences between distributions of correct and incorrect generations on an idiom dataset, and we observe that this output cross-entropy appears to correspond with model uncertainty given a prompt. 

Future work will aim to address limitations of this preliminary study by expanding our analysis to a broad range of datasets and language models of varying sizes. We will additionally extend our study to multi-token generations and explore the use of output cross-entropy as an uncertainty measure and potential flag for hallucinations. Finally, we intend to explore additional convergence metrics that may better predict correct versus incorrect generations, then examine how broadly applicable they are across models and datasets. The ultimate goal of this research is to develop methods for assuring the quality of language model output with minimal computational cost.

\begin{ack}
We thank Cash Costello and Patrick Emmanuel for providing helpful feedback. Funding was provided via an internal research grant from Johns Hopkins University Applied Physics Lab. Compute hours were provided on Oak Ridge National Laboratory's Summit supercomputer as a part of an award for this project from the National Science Foundation's National Artificial Intelligence Research Resources (NAIRR) Pilot Program. 
\end{ack}

\bibliographystyle{plain}
\bibliography{neurips_2024}{}

\appendix

\section{Appendix / supplemental material}

\subsection{Cross-Entropy Measurement}
\label{sec:appendix-cross-entropy}
Given two discrete probability distributions $p$ and $q$ over support $\mathcal{X}$, the cross-entropy $\mathcal{H}$ of $q$ relative to $p$ is defined as
\[
    \mathcal{H}(p,q) = - \sum_{x \in \mathcal{X}} p(x)\ log\ q(x)
\]

In our case, $\mathcal{X}$ represents the vocabulary of tokens, $q$ represents the probabilities predicted by the model over $\mathcal{X}$, and $p$ represents some target distribution over $\mathcal{X}$. For our methods, we take $p$ to be a one-hot encoding representing either the sampled token $\hat{y}$ or the ground truth token $y$. With a one-hot encoding for $p$, the above summation yields only one non-zero term

\[
    \mathcal{H}(p,q) = -log\ q(x^*)
\]

where $x^* = y$ or $\hat{y}$ depending on our choice of target. This is also the negative log likelihood of token $x^*$ as predicted by the model. Our cross-entropy experiments can be equivalently framed as measuring how the negative log likelihood of a target token changes throughout the inference process.

\subsection{Best and Worst Case Sample Generations}
\label{sec:appendix-regret-samples}

\begin{figure}[!ht]
  \centering
  \subfloat{\includegraphics[width=1\textwidth]{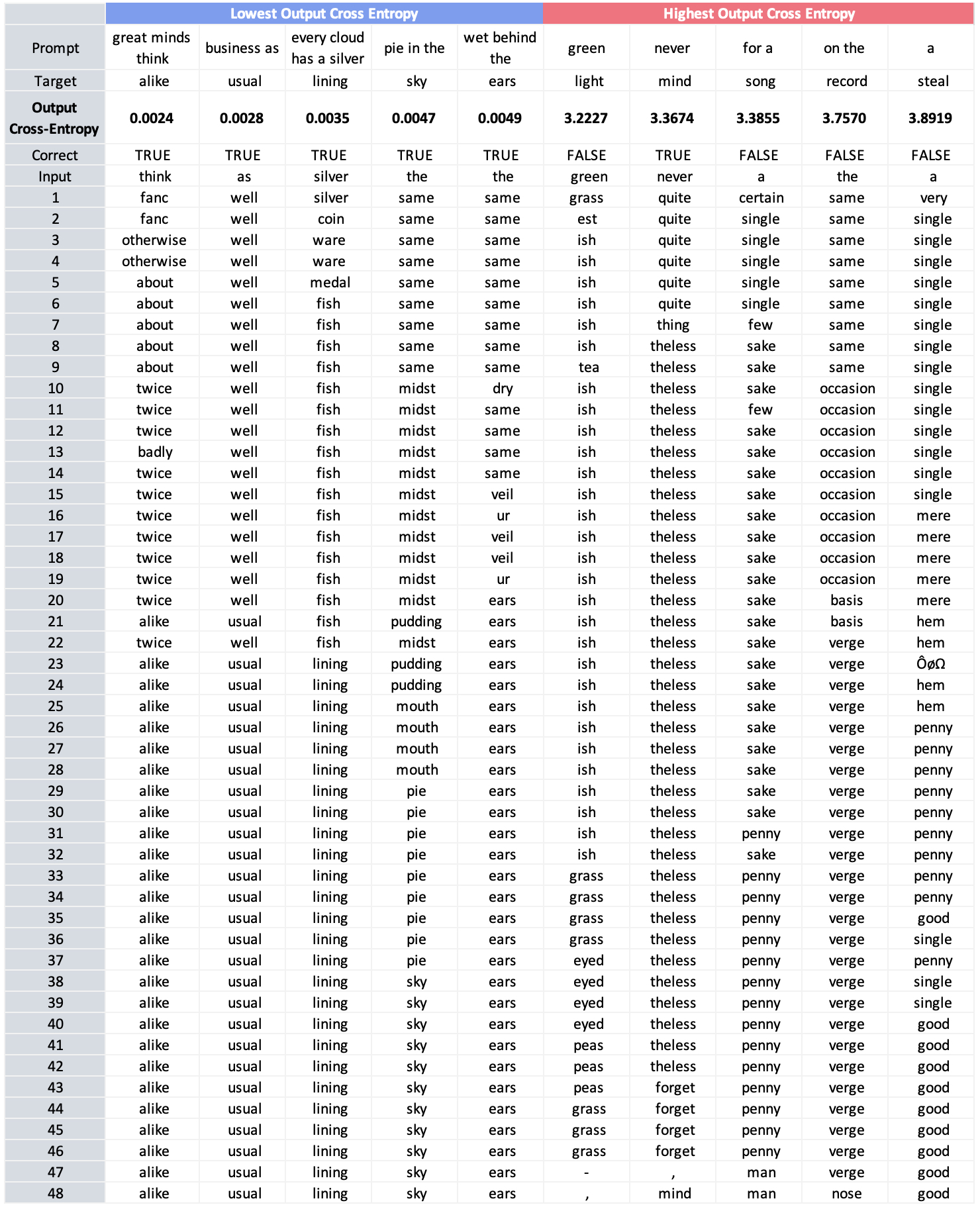}}
  \caption{A look into the residual stream for the idiom generations with highest and lowest output cross-entropy. The token corresponding to the highest logit in the residual prediction after each layer is displayed to show how the path through token space.}
\label{idiom-table}
\end{figure}

We present the residual predictions for the generations with the highest and lowest output cross-entropy scores vs model predictions $\hat{y}$ on the idiom dataset in Figure~\ref{idiom-table}. These correspond to left- and right-most samples on the cross-entropy axis in Figure~\ref{idiom-output-cross-entropy}a. We use the logit lens technique to recover token predictions from the residual stream after each layer update as described in \ref{sec:methods}. 

Many of the idiom prompts are one word and are extremely difficult to complete, even for humans. This open-endedness lends itself to high uncertainty, which is captured by the cross-entropy as shown here. The samples with the lowest cross-entropy have prompts with multiple words that heavily imply a specific next-token, severely constraining the set of possible next-tokens. In contrast, samples with the highest cross-entropy have short prompts with very common words that are could have many valid next tokens, resulting in a very open-ended generation task. We observe output cross-entropy reflecting the open-endedness of the prompt via the corresponding uncertainty from the model. 

\subsection{Choice of Divergence Target}
\label{sec:appendix-output-targets} 

\begin{figure}[!ht]
  \centering
  \subfloat[][KL Divergence (\textit{layer}, $\hat{y}$ \textit{logits})]{\includegraphics[width=.485\textwidth]{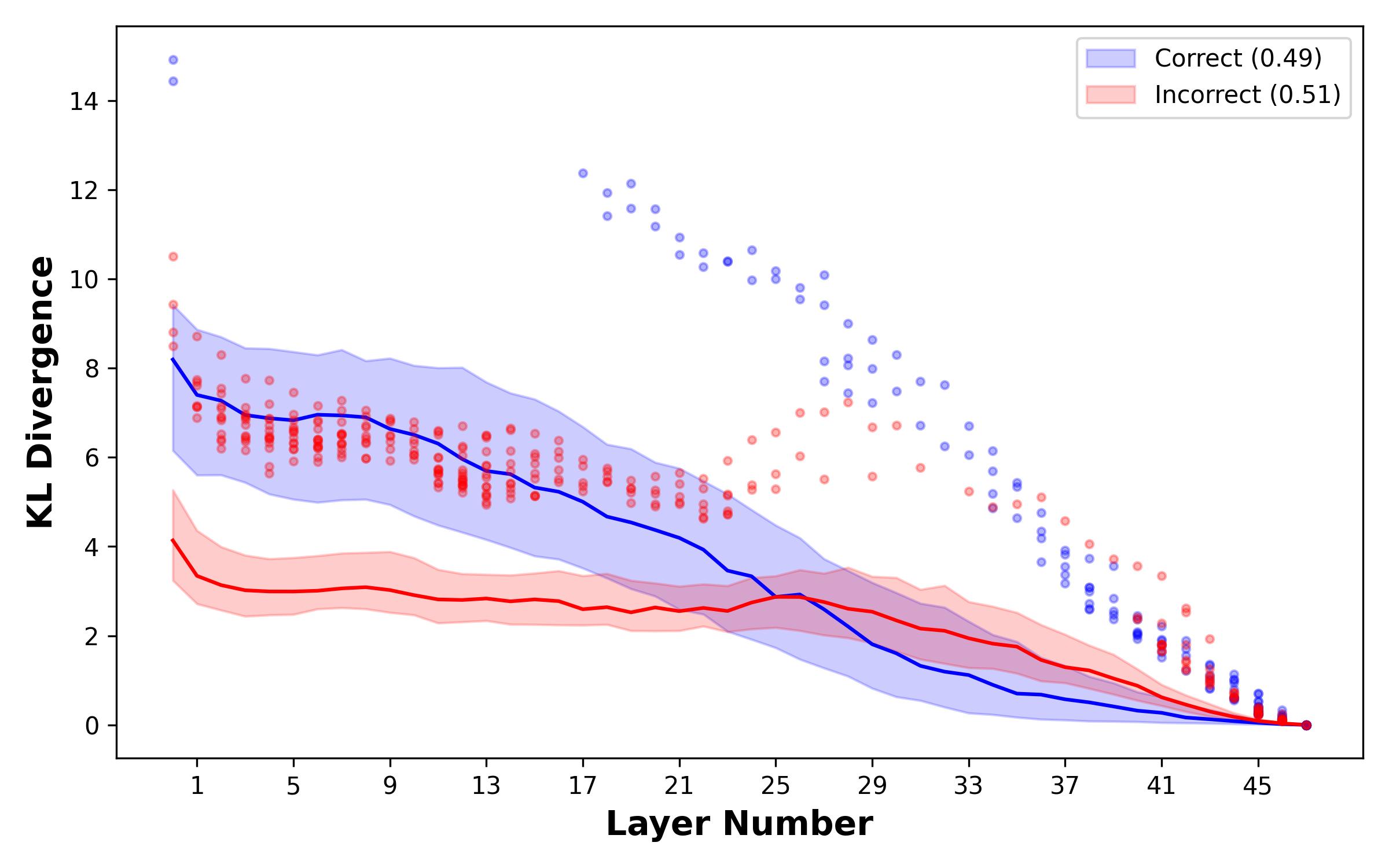}}\quad
  \subfloat[][KL Divergence (\textit{layer}, $\hat{y}$ \textit{one-hot})]{\includegraphics[width=.485\textwidth]{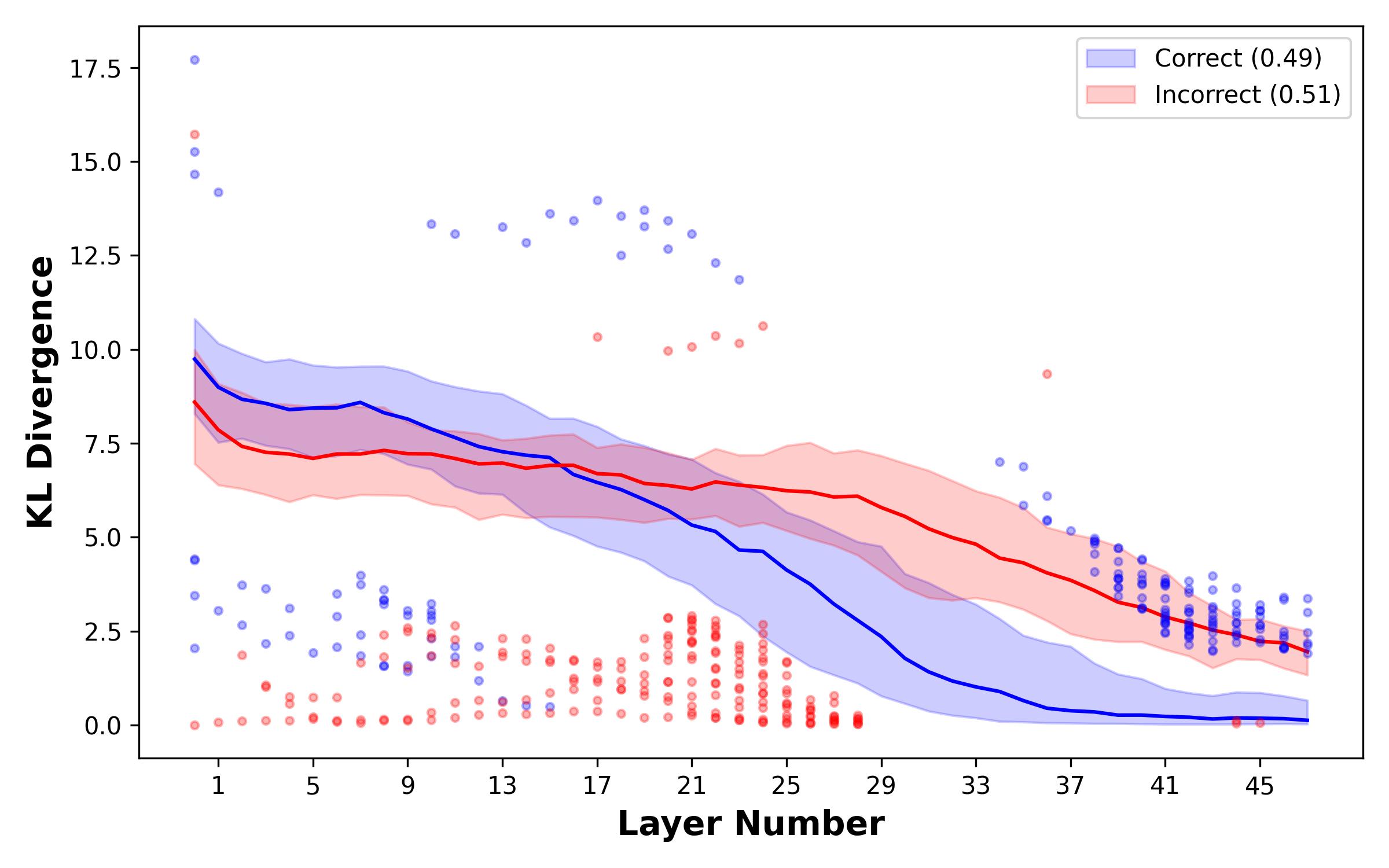}}
  \caption{ Computing KL divergence with respect to (a) the model output logits versus (b) the one-hot distribution representing the top logit in the model output. It can be observed that (b) implicitly penalizes outputs with high entropy. }
\label{idiom-output-targets}
\end{figure}

Here we compare KL divergence between residual predictions and model output logits versus residual predictions and a one-hot encoding of the top predicted logit. Note that KL divergence and cross-entropy between two distributions differs only by a constant. By definition, the final residual prediction is equivalent to the model output logits, thus the KL divergence approaches zero for all generations, as observed in Figure~\ref{idiom-output-targets}a below. If the output logits of the model exhibit high entropy, then they will have a higher divergence when measured against to a one-hot representation of the top predicted logit. This can be observed in Figure~\ref{idiom-output-targets}b. We find this bias useful for distinguishing correct and incorrect generations. 

\subsection{Idiom Dataset Loss Table}
\label{sec:delta-loss}

Figure~\ref{idiom-delta-loss} below shows average change in loss after each layer update to predictions in the residual stream for the idiom dataset. White cells denote no change, blue cells denote a decrease, and red cells denote an increase. Nearly all layer updates move the residual prediction in a direction of decreasing loss, supporting the IIH. Layers 20-38 appear to contribute the most to reducing residual prediction loss for correct generations. 

\begin{figure}[!ht]
  \centering
  \subfloat{\includegraphics[width=1.0\textwidth]{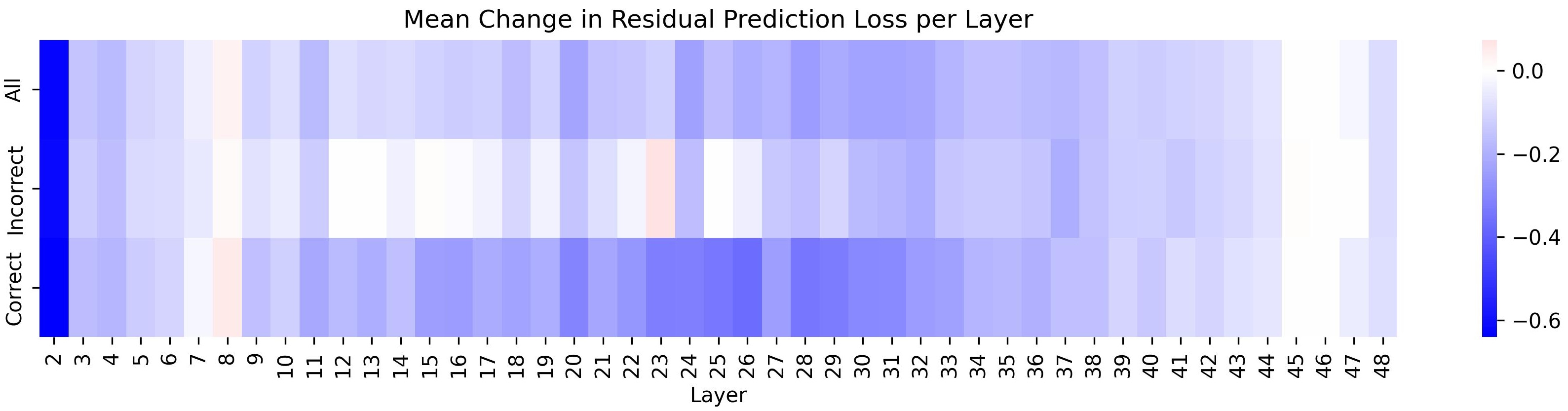}}
  \caption{The change in cross-entropy loss between the prediction in the residual stream and the ground truth $y$ after each layer update, averaged across all prompts in the idiom dataset. Nearly all updates reduce the loss on average across all groupings. }
\label{idiom-delta-loss}
\end{figure}

\subsection{Idiom Dataset Cosine Similarity}

We also measure how representations change directly in token embedding space by measuring the cosine similarity between each intermediate embedding $e_n^i$ and embeddings of the two choices of targets described in \ref{sec:cross-entropy}. 

\begin{figure}[!ht]
  \centering
  \subfloat[][Cosine Similarity (layer, $\hat{y}$)]{\includegraphics[width=.485\textwidth]{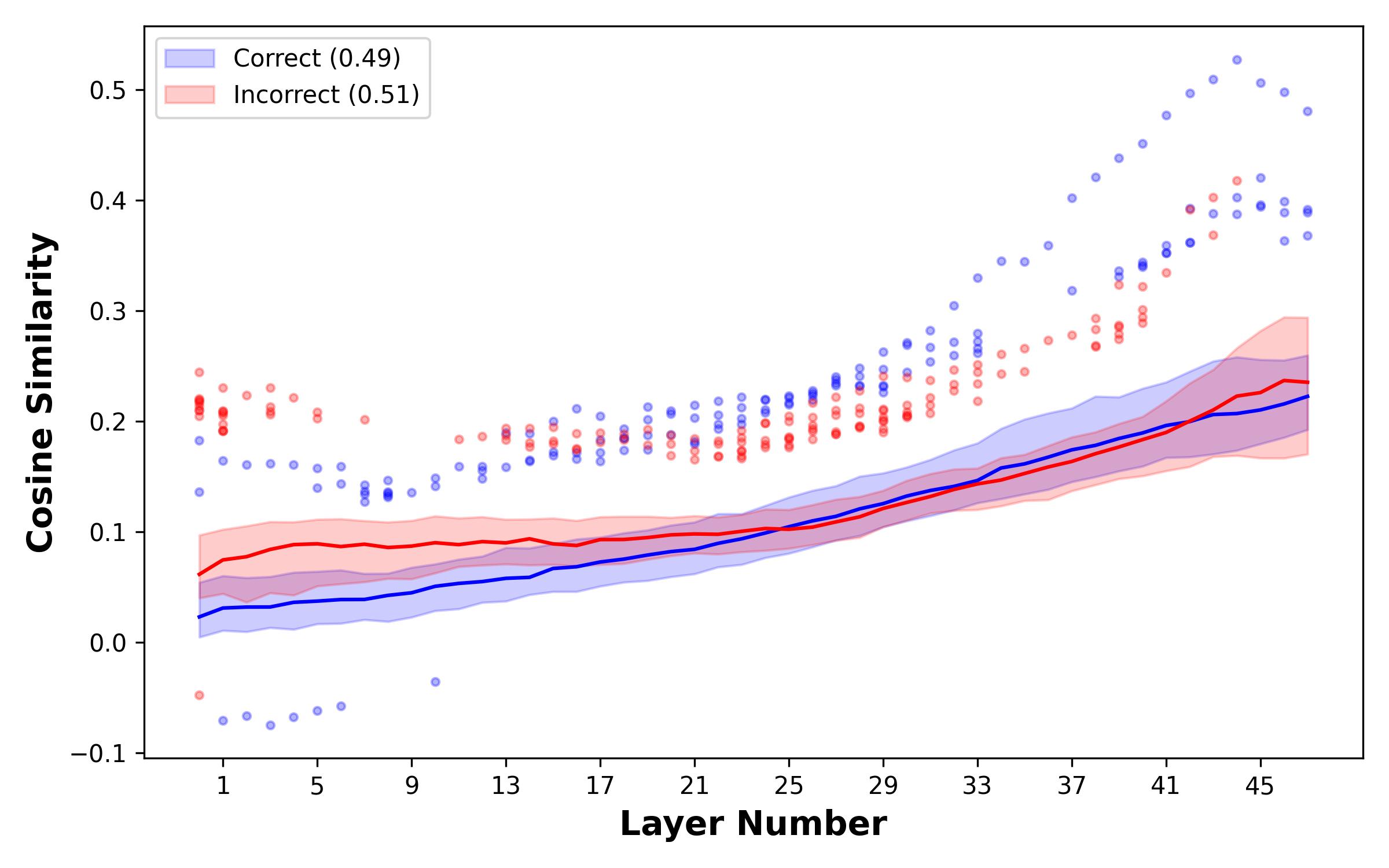}}\quad
  \subfloat[][Cosine Similarity (layer, $y$)]{\includegraphics[width=.485\textwidth]{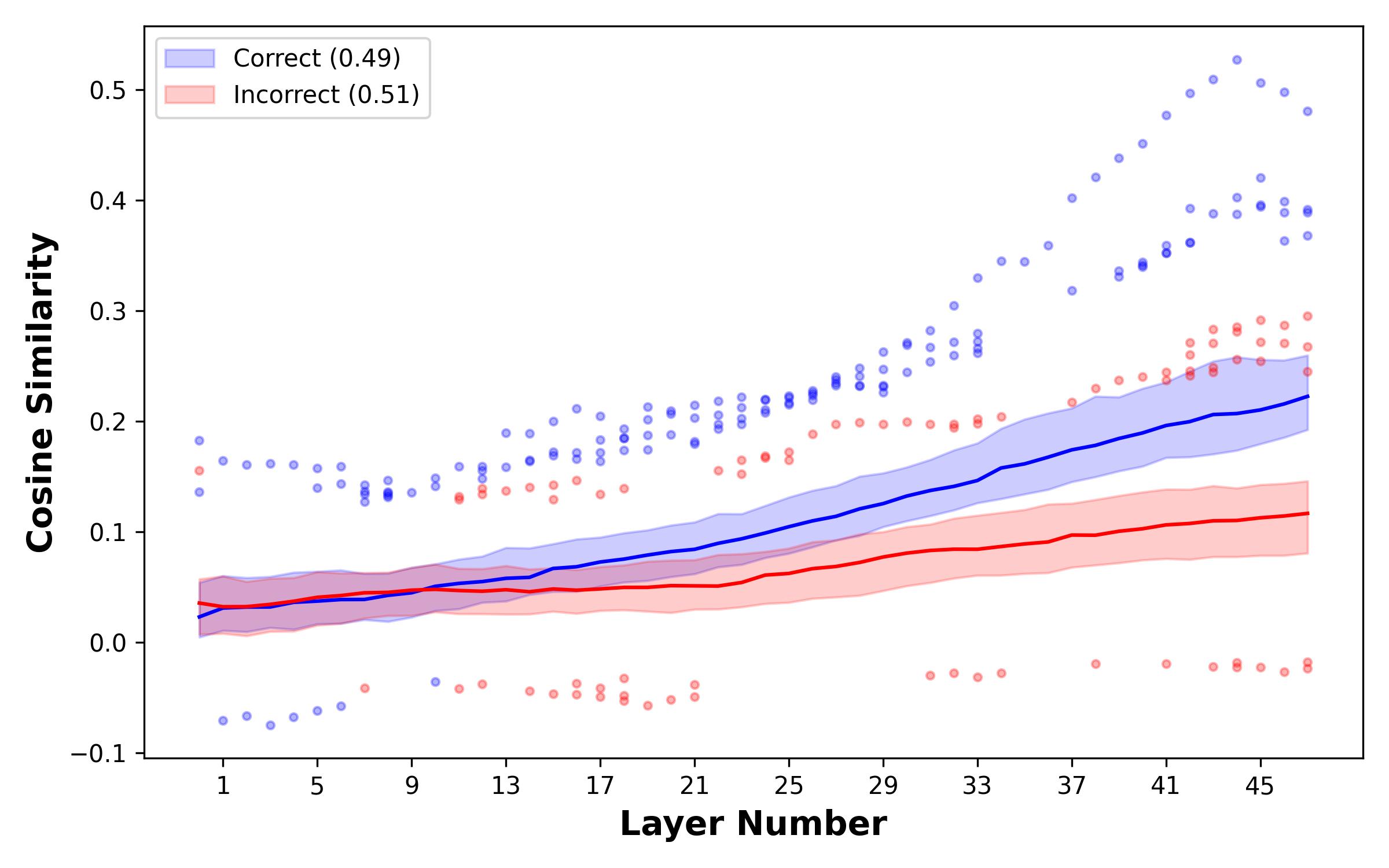}}  
  \caption{GPT-2 iteratively increases cosine between the embedding of the \(n^{th}\) token in the residual stream and the most-likely next token. The median, inter-quartile ranges, and outliers of correct and incorrect generations are plotted in blue and red. }
\label{idiom-cosine}
\end{figure}

\end{document}